%% file: iclr2027_conference.tex
\documentclass{article} 
\usepackage{iclr2027_conference,times}

\input{math_commands.tex}

\usepackage{hyperref}
\usepackage{url}
\usepackage{booktabs}
\usepackage{cleveref}
\usepackage{bm}
\usepackage{graphicx}
\usepackage{subcaption}
\usepackage{multirow}
\usepackage{wrapfig}

\title{Reading the Whole Heart: Latent-Attention Masked Autoencoders for Multimodal \\Cardiac Representation Learning}

\author{%
\normalfont%
Andrea Agostini\textsuperscript{1,*},
Simon Böhi\textsuperscript{2,*},
Moritz Vandenhirtz\textsuperscript{1},
Samuel Ruiperez-Campillo\textsuperscript{1},
\\
Max Krähenmann\textsuperscript{2},
Silke Mühlstedt\textsuperscript{1},
Irene Cannistraci\textsuperscript{1},
\\
Ece Özkan Elsen\textsuperscript{2,$\dagger$},
Julia E. Vogt\textsuperscript{1,$\dagger$},
Thomas M. Sutter\textsuperscript{1,$\dagger$}
\\[0.7em]
\textsuperscript{1}Department of Computer Science, ETH Zurich, Switzerland \\
\textsuperscript{2}Department of Biomedical Engineering, University of Basel, Switzerland
\\[0.5em]
\textsuperscript{*}Equal contribution.
\quad
\textsuperscript{$\dagger$}Shared senior authorship.
}

\iclrfinalcopy 
\begin{document}

\maketitle
\begin{abstract}
Cardiovascular diagnosis and treatment rest on integrating complementary modalities, such as electrocardiogram, echocardiography, and chest X-rays, each capturing distinct but complementary aspects of cardiac pathophysiology.
Yet most medical foundation models remain modality-specific, combining modalities only for finetuning or post-training.
This discards the cross-modal evidence clinicians naturally integrate and ignores the structure within each modality.
We introduce Latent-Attention Masked Autoencoder (LAMAE), a multimodal, structure-aware masked autoencoder that jointly learns patient-level representations during self-supervised pretraining.
Instead of fusing modalities post hoc, LAMAE exchanges information directly in the latent space through a shared latent-attention module operating over a study–view–entity hierarchy, enabling aggregation of variable observations and handling of missing modalities.
Pretrained on over $500{,}000$ MIMIC-IV hospital stays, LAMAE outperforms modality-specific pretraining and strong contrastive and vision–language baselines across multimodal hospital-stay tasks, such as in-hospital mortality, ICD-10 and DRG coding, and length of stay, while remaining competitive on unimodal tasks.
Modeling this structure also pays off within a single modality: even without cross-modal information, the latent-attention module improves representations over modality-specific pretraining.
\end{abstract}

\input{01_intro}
\input{02_related_work}
\input{03_method}
\input{04_experiments}
\input{05_conclusion}

\section*{Acknowledgments}
AA is supported by the Swiss AI project ``Beyond Signals and Structure: Enabling Generalizable Multimodal Cardiopulmonary AI’’ under contract number SNAI000046. MV is supported by the Swiss State Secretariat for Education, Research and Innovation (SERI) under contract number MB22.00047. This work was supported by computational resources provided by the Swiss National Supercomputing Centre (CSCS) under the Alps infrastructure through the Swiss AI Initiative (project ID aa012).

\bibliography{iclr2027_conference}
\bibliographystyle{iclr2027_conference}

\newpage
\appendix
\input{06_appendix}

\end{document}

%% file: math_commands.tex
\usepackage{amsmath,amsfonts,bm}

\def\eqref#1{equation~\ref{#1}}

\def\1{\bm{1}}

\DeclareMathAlphabet{\mathsfit}{\encodingdefault}{\sfdefault}{m}{sl}
\SetMathAlphabet{\mathsfit}{bold}{\encodingdefault}{\sfdefault}{bx}{n}



%% file: 01_intro.tex
\section{Introduction}
\label{sec:intro}
Cardiovascular diseases (CVDs) remain the leading cause of mortality worldwide, accounting for more than 20 million deaths annually and imposing a substantial medical and financial burden on healthcare systems \citep{global2025global}.
Thus, early diagnosis and appropriate risk stratification are critical for improving patient outcomes. Notably,  clinical decision making rarely relies on a single diagnostic modality \citep{Acosta2022}.
Instead, clinicians routinely integrate complementary information from electrocardiograms (ECGs), echocardiography (Echo), chest radiographs (CXR), laboratory values, and clinical history to form a comprehensive assessment of cardiovascular diagnosis and prognosis \citep{kashou2023ecg,dohi_echocardiographic_2019}.

Recent advances in self-supervised learning (SSL) and foundation models have demonstrated that large-scale pretraining can learn transferable representations from unlabeled medical data, thereby substantially reducing the dependence on expensive expert annotations \citep{azizi_big_2021,khan2025comprehensive}.
Masked autoencoders \citep[MAEs,][]{he2022masked} have emerged as a scalable paradigm for representation learning by reconstructing masked portions of the input from sparse observations.
MAEs have since been successfully adapted to numerous medical domains, including CXRs \citep{xiao_delving_2023}, Echocardiography \citep{jiao_usfm_2024,kim2025echofmfoundationmodelgeneralizable}, and ECG \citep{na2024guiding,jin2025reading}.
Despite these advances, nearly all existing foundation models, and especially their pretraining strategies, remain modality-specific, learning separate representations for ECG, Echo, or CXR independently before combining them.
If at all, different modalities are only combined during downstream prediction.

Beyond being modality-specific, many existing approaches also overlook the intrinsic structure present within medical data.
Medical observations are rarely independent.
A standard 12-lead ECG consists of multiple complementary voltage vector projections of the same cardiac electrical activity, while CXR and Echo examinations comprise multiple anatomical views acquired from different imaging angles.
These findings suggest that exploiting structural relationships constitutes an effective form of inductive bias for medical foundation models.
However, current approaches remain confined to individual modalities and therefore fail to exploit the relationships between complementary diagnostic modalities acquired from the same patient.
In current clinical practice, clinicians take advantage of these diagnostic modalities for many clinically important diseases, including acute myocardial infarction, heart failure, valvular heart disease, and cardiomyopathies \citep{mcdonagh_2021_2021,byrne_2023_2023}.
Each modality provides complementary evidence, which clinicians naturally integrate for diagnosis.

However, learning from heterogeneous diagnostic modalities presents several challenges:
First, different modalities exhibit different data characteristics, ranging from one-dimensional physiological time series to two-dimensional images and spatio-temporal videos.
Second, clinical examinations are inherently incomplete: patients often undergo only a subset of available diagnostic procedures, requiring models to handle arbitrary patterns of missing modalities.
Finally, multimodal observations exhibit hierarchical structure, with dependencies both within modalities (e.g., ECG leads or echocardiographic views) and across modalities at the patient level.
Existing multimodal learning approaches typically rely on contrastive alignment or late feature fusion, limiting their ability to directly model these hierarchical interactions during representation learning.

In this work, we propose Latent-Attention Masked Autoencoder (LAMAE), a multimodal and structure-aware MAE \citep{he2022masked} that jointly learns representations from ECG, Echo, and CXRs through a shared latent attention (LA) mechanism.
Rather than treating modalities as independent feature extractors whose representations are fused only during downstream prediction, our LAMAE enables information exchange directly within the latent space during self-supervised pretraining.
This design naturally accommodates heterogeneous modalities, variable numbers of observations, and missing data while preserving the flexibility to exploit modality-specific information when appropriate. 
Our framework learns unified patient-level representations that better reflect the multimodal nature of clinical reasoning.

Our contributions are three-fold:
(1) We introduce Latent-Attention Masked Autoencoders (LAMAE), a structure-aware multimodal masked autoencoder for ECG, Echo, and CXRs: a shared latent-attention (LA) module exchanges information directly in the latent space over a study--view--entity hierarchy, paired with an entity-level masking scheme that drops complete leads, frames, or views to force reconstruction across modalities and views, as illustrated in Figure~\ref{fig:model}.
(2) We show that modeling this structure improves representations at every level: LAMAE outperforms modality-specific pretraining and strong contrastive and vision--language baselines on hospital-stay-level tasks, and the LA module improves representations even within a single modality, without any cross-modal information.
(3) We show that multimodal pretraining outperforms unimodal pretraining on hospital-stay tasks without sacrificing unimodal performance.

%% file: 02_related_work.tex
\section{Related Work}
\label{sec:related_work}

\noindent
\textbf{Self-supervised pretraining in medicine.}
Self-supervised learning has become a key strategy for medical representation learning, where expert annotation is costly and labels are imbalanced \citep{azizi_big_2021, moody2025foundation}.
Contrastive frameworks \citep{chen_simple_2020, he2020momentum} applied to large corpora such as MIMIC-CXR \citep{johnson2019mimic} yield encoders that transfer well \citep{azizi_big_2021, tiu_expert-level_2022}. However, they treat every image as an independent instance.
Masked autoencoders (MAEs) reconstruct masked patches \citep{he2022masked, bizeul_pixels_2025} or spatiotemporal tubes \citep{tong_videomae_2022, wang_videomae_2023}. They have been adapted separately to CXR \citep{xiao_delving_2023}, Echo \citep{jiao_usfm_2024, kim2025echofmfoundationmodelgeneralizable, zhang_echo-vision-fm_2024, yang_echocardmae_2026} and ECG \citep{na2024guiding, jin2025reading}.
The resulting encoders are siloed, and their reconstruction signal never leaves a single modality.

\noindent
\textbf{Multimodal medical models.}
Vision--language models pair images with reports \citep{boecking2022making, chen2024chexagent, deperrois2025radvlm}, generalist models extend this to several imaging domains \citep{sellergren_medgemma_2026, zhang_multimodal_2026, gao_probmed_2025}, and ImageBind \citep{girdhar_imagebind_2023} aligns many modalities to a single anchor.
In cardiology, Echo2ECG \citep{liman_echo2ecg_2026} contrastively aligns ECGs with multi-view echocardiograms.
Text-anchored supervision is semantic rather than physiological: a report states what is seen, not how the different recordings of the same heart relate to each other.
Contrastive alignment, whether to text or between modalities, additionally requires paired samples, which are scarce for many modality combinations in routine care (\cref{fig:modality_upset}).

\noindent
\textbf{Multimodal masked autoencoders.}
MultiMAE \citep{bachmann_multimae_2022} and 4M \citep{mizrahi_4m_2023} reconstruct masked tokens of several modalities with a single shared encoder, and M3AE \citep{geng_multimodal_2022} does the same for image--text pairs.
\citet{turgut_unlocking_2025} combine masked data modeling with contrastive alignment to transfer information from cardiac MRI to the ECG, but this requires paired studies of two modalities.
LAMAE instead keeps modality-specific encoders at the entity level and fuses modalities only through a shared latent-attention module. This lets it pretrain on any observed subset of modalities.

\noindent
\textbf{Structured studies.}
Medical data are rarely single-entity samples: a CXR study couples frontal and lateral projections, an ECG comprises twelve simultaneous leads, and an echocardiographic exam consists of several videos from complementary views.
Prior work exploits this through multi-view or correspondence-based contrast \citep{sutter2024unity, agostini2025leveraging, laguna2025structure}, hierarchical fusion \citep{mokhtari_gemtrans_2023}, latent-space MAEs over frozen per-view encoders \citep{tohyama_multi-view_2025}, or lead alignment combined with masked reconstruction \citep{erlacher2025swissbeatsnet}. These approaches mostly rely on \emph{pairwise} objectives.
LAMAE instead provides a unified framework for modeling structure both within and across modalities, jointly integrating studies, views, and entities through latent attention.
In this way, LAMAE learns not from \emph{more} data, but from the \emph{structure} of the data itself.

%% file: 03_method.tex
\section{Method}
\label{sec:method}
\begin{figure}[t]
    \centering
    \includegraphics[width=0.9\textwidth]{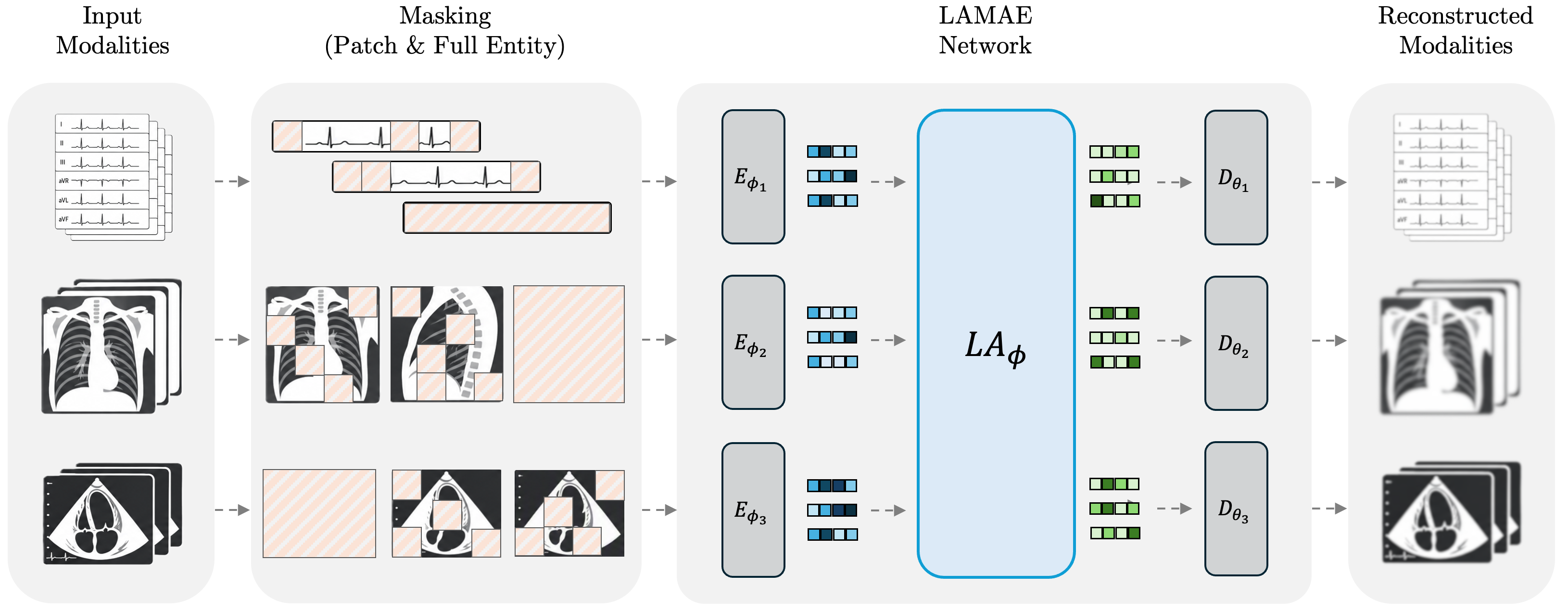}
    \caption{\textbf{Overview of the LAMAE framework.}
From left to right: \textit{(1) Input modalities:} a sample contains a variable set of ECG, CXR, and Echo observations.
\textit{(2) Masking:} inputs are masked at both the patch and full-entity levels, with entity masking removing complete entities such as ECG leads, CXR views, or Echo frames.
\textit{(3) LAMAE network:} modality-specific encoders $E_{\phi_k}$ independently encode the visible tokens, after which the shared latent attention module $LA_\phi$ jointly processes their latent representations, enabling information exchange within and across modalities. Modality-specific decoders $D_{\theta_k}$ then reconstruct the masked inputs.
\textit{(4) Reconstructed modalities:} the reconstructed ECG, CXR, and Echo observations provide the self-supervised pretraining objective.}
    \label{fig:model}
    \vspace{-0.5cm}
\end{figure}

We propose Latent-Attention Masked Autoencoders (LAMAEs), a multimodal and structure-aware MAE \citep{he2022masked} that jointly learns representations from multimodal datasets $\mathbb{S}$ (\Cref{fig:model}).
We define a multimodal dataset as $\mathbb{S} = \{ \mathcal{S}^{(s)} \}_{s=1}^{S}$ where every multimodal sample $\mathcal{S}^{(s)} = \left( \{\bm{X}^{(s,k)}\}_{k\in\mathcal{M}}, \bm{m}^{(s)} \right)$ pairs the per-modality data $\bm{X}^{(s,k)}$, $k \in \mathcal{M}$, with an availability mask $\bm{m}^{(s)} \in \{0, 1 \}^{|\mathcal{M}|} $ whose entries are
\begin{align}
    m^{(s,k)}=
\begin{cases}
1,&\text{if modality }k\text{ is observed for }s,\\
0,&\text{otherwise.}
\end{cases}.
\end{align}
The observed modalities per multimodal sample $s$ are then $\mathcal{M}_{\text{obs}}^{(s)} = \{ k \in\mathcal{M}\mid m^{(s,k)}=1\}$.

\begin{figure}[h]
    \centering
    \includegraphics[width=0.9\textwidth]{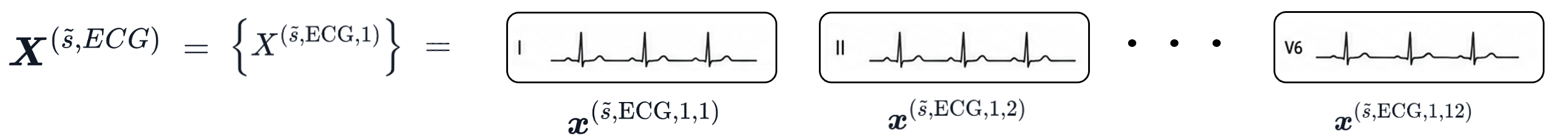}
    \caption{\textbf{Example of the study-view-entity hierarchy for ECG.} For an example sample $\tilde{s}$, the ECG modality $\bm{X}^{(\tilde{s},\mathrm{ECG})}$ consists of a single ECG examination $X^{(\tilde{s},\mathrm{ECG},1)}$ ($v=1$), which comprises $E_{\mathrm{ECG}}=12$ base entities $\bm{x}^{(\tilde{s},\mathrm{ECG},1,e)}$, corresponding to the 12 ECG leads. Leads I, II, and V6 are shown for illustration.}
    \label{fig:example_ecg}
    \vspace{-0.5cm}
\end{figure}

The data acquisition process of healthcare data leads to more structure in the data compared to general or natural domain data \citep{laguna2025structure}.
Therefore, we introduce a three-layered hierarchy to describe modalities: \textit{studies}, \textit{views}, \textit{and base entities}.
Every modality $\bm{X}^{(s,k)}$ is a medical study that may consist of multiple examinations, e.g., an Echo study usually consists of multiple videos of different frames each.
Hence, a modality is given as $\bm{X}^{(s,k)} = \{ X^{(s,k,1)}, \ldots, X^{(s,k,V_k)} \}$, where $X^{(s,k,v)}$ is the data sample for view $v$ and modality $k$, e.g., video of an echo study.
For every modality $k$ and sample $s$, we define the set of observed views as $\mathcal{V}_{k,\text{obs}}^{(s)} \subseteq \mathcal{V}_k = \{1, \ldots, V_k \}$, where $V_k = |\mathcal{V}_k|$ is the total number of views and $\mathcal{V}_k$ the complete set of views for modality $k$.
Every view $v$ of modality $k$ is defined as $X^{(s,k,v)} = (\bm{x}^{(s,k,v,1)}, \ldots, \bm{x}^{(s,k,v,E_k)})$, where each base entity $\bm{x}^{(s,k,v,e)}$ is, e.g., a frame in an Echo video, a lead in an ECG measurement, or an image in a CXR study, and $\mathcal{E}_k = \{ 1, \ldots, E_k \}$ is the set of base entities with $E_k = |\mathcal{E}_k|$. We illustrate this hierarchy for an ECG study in \Cref{fig:example_ecg}.

\subsection{Encoders and Decoders}
\label{sec:method_enc_dec}
We have modality-specific encoders $E_{\phi_k}(\cdot)$ and decoders $D_{\theta_k}(\cdot)$.
Following \citet{he2022masked}, we use transformer-based encoder and decoder architectures \citep{vaswani_attention_2017}.
More specifically we follow the suggested model variants presented in \citet{dosovitskiy_image_2020} and \citet{he2022masked}.
For details regarding the modality-specific adaptations, we refer to \cref{app:implementation}.

As it is typical for MAEs, we mask a large portion of the input $\bm{X}^{(s,k)}$ during pretraining with the objective to reconstruct the masked input.
Given a binary mask $M_{k,\alpha_k}^{(s)}(\cdot)$ with mask ratio $\alpha_k$ for modality $k$, we define the model input  $\bm{X}_{vis}^{(s,k)}$ as
\begin{align}
\label{eq:masking_basic}
    \bm{X}_{vis}^{(s,k)} = M_{k,\alpha_k}^{(s)} (\bm{X}^{(s,k)}).
\end{align}

During pretraining, the encoder $E_{\phi_k}$ only processes the visible part $\bm{X}_{vis}^{(s,k)}$ and embeds them into latent embeddings $\bm{Z}^{(s,k)}$ to reconstruct the input $\hat{\bm{X}}^{(s,k)}$.
For every modality $k$, the encoder $E_{\phi_k}$ sequentially processes modalities on a base entity level $\bm{x}^{(s,k,v,e)}$.
For every $\bm{x}^{(s,k,v,e)}$, we have $T_k$ patches and $T_{k,vis} = (1 - \alpha_k) \cdot T_k$ visible patches.
Hence, $\mathcal{T}_{k,vis}^{(s)} \subseteq  \{1, \ldots, T_k \}$ is the set of input patches for entity $e$ of modality $k$.
We then have
\begin{align}
\label{eq:mm_enc}
    \bm{z}^{(s,k,v,e)} = E_{\phi_k}(\bm{x}^{(s,k,v,e)}) \hspace{0.25cm} \text{such that} \hspace{0.25cm} \bm{Z}^{(s,k)} = \{ \bm{z}^{(s,k,v,e)} \mid e \in \mathcal{E}_k, v \in \mathcal{V}_{k,\text{obs}}^{(s)}\}
\end{align}
Following standard practice in ViTs \citep{dosovitskiy_image_2020}, each latent embedding $\bm{z}^{(s,k,v,e)}$ consists of the latent embeddings $\bm{z}_{t}^{(s,k,v,e)}$ of the visible patches $t \in \mathcal{T}_{k,vis}^{(s)}$ and an entity-level CLS embedding $\bm{z}_{\mathrm{cls}}^{(s,k,v,e)}$.
For simplicity, we describe the set of latent embeddings $\bm{Z}^{(s,k)}$ for modality $k$ as $\bm{Z}^{(s,k)} = E_{\phi_k} (\bm{X}_{vis}^{(s,k)})$.
Similarly, we have modality-specific decoders $D_{\theta_k}$, which decode the base entities of each modality separately, i.e.,
\begin{align}
\label{eq:mm_dec}
    \hat{\bm{x}}^{(s,k,v,e)} = D_{\theta_k}(\bm{z}_{out}^{(s,k,v,e)}) \hspace{0.25cm} \text{where} \hspace{0.25cm} \bm{Z}_{out}^{(s,k)} = \{ \bm{z}_{out}^{(s,k,v,e)} \mid e \in \mathcal{E}_k, v \in \mathcal{V}_{k,\text{obs}}^{(s)} \}.
\end{align}
For vanilla MAEs, we have $\bm{z}^{(s,k,v,e)} = \bm{z}_{out}^{(s,k,v,e)}$.

\subsection{Latent Attention for Multimodal Integration}
Multimodal datasets $\mathbb{S}$ provide additional structure in the data acquisition process that we can leverage.
We introduce a novel \emph{Latent Attention} (LA) module for leveraging the additional structure in the data.
Inspired by \citet{ilse2018attention} and \citet{lee2019set}, the LA module
allows to learn the correlation and shared information between different modalities, views or entities.
The LA module is where the information coming from all modalities is being processed jointly.
Let $\bm{Z}^{(s)} = \{\bm{Z}^{(s,k)} \}_{k \in \mathcal{M}_{\text{obs}}^{(s)}}$ be the concatenation of latent embeddings $\bm{Z}^{(s,k)}$ of all modalities $k \in \mathcal{M}_{\text{obs}}^{(s)}$.
Given a set of latent embeddings $\bm{Z}^{(s)}$, the LA module transforms the embeddings $\bm{z}^{(s,k,v,e)}$ that only contain information of their respective base entities into embeddings containing all available multimodal information as follows
\begin{align}
    \bm{Z}_{out}^{(s)} = LA_\phi(\bm{Z}^{(s)}) = \{ \bm{z}_{out}^{(s,k,v,e)} \mid k \in \mathcal{M}_{\text{obs}}^{(s)}, v \in \mathcal{V}_{k,\text{obs}}^{(s)}, e \in \mathcal{E}_k \}.
\end{align}
We design the LA module as a multi-head self-attention block that attends jointly over the patch and CLS embeddings of all observed entities, views and modalities (see \Cref{app:implementation}).

\subsection{Entity Masking}
The LA module allows for more elaborate masking strategies compared to single-entity-based approaches, such as the independent (or vanilla) MAE \citep{he2022masked}.
Independent MAEs define their pretraining objective via patch-based masking: a signal or image is divided into patches, of which a random subset is masked, defining the reconstruction target (see \cref{sec:lamae} for details).
The LA module additionally allows us to mask or drop complete entities such as leads, frames, or views.
Dropping complete entities forces the LA module to connect information across base entities.
 
We make \Cref{eq:masking_basic} more specific as follows. Let $\mathcal{E}_k$ denote the entity set of modality $k$, with $E_k = |\mathcal{E}_k|$ assumed constant across the views $v$ of modality $k$, and assume all entities of a given $(k,v)$ carry the same number of tokens. For each sample $s$, modality $k$, and view $v$, we independently sample the number of visible (encoded) entities
\begin{align}
    \mathcal{E}_{vis}^{(s,k,v)} \subseteq \mathcal{E}_k
    \quad\text{with}\quad
    \big|\mathcal{E}_{vis}^{(s,k,v)}\big| = n_{e,vis}^{(s,k,v)}
    \sim \mathcal{U}\{n_{\min}, n_{\max}\},
\end{align}
and treat the remaining $E_k - n_{e,vis}^{(s,k,v)}$ dropped entities as fully masked
(mask ratio $100\%$).
The reconstruction target comprises all masked tokens, i.e., every token of a dropped entity together with the masked tokens of the visible entities.
 
We target a mask ratio $\alpha_k$ per modality, held constant across all views $v$ and all samples $s$. Since $n_{e,vis}^{(s,k,v)}$ varies across views, we set the per-entity mask ratio $\beta^{(s,k,v)}$ applied to the visible entities so that each view attains the common target $\alpha_k$ in expectation.
Requiring $\big(E_k - n_{e,vis}^{(s,k,v)} + n_{e,vis}^{(s,k,v)}\beta^{(s,k,v)}\big)/E_k = \alpha_k$
gives
\begin{align}
    \beta^{(s,k,v)}
    = \frac{\alpha_k - 1 + n_{e,vis}^{(s,k,v)}/E_k}{\,n_{e,vis}^{(s,k,v)}/E_k\,}.
\end{align}
For $\beta^{(s,k,v)} \geq 0$ the number of dropped entities must not by itself exceed the target, which bounds the sampler via $n_{\min} \geq \lceil (1-\alpha_k)\,E_k \rceil$.
 
The resulting masking procedure is
\begin{align}
    \bm{X}_{vis}^{(s,k)}
    = \big\{\, M_{k,\beta^{(s,k,v)}}(\bm{x}^{(s,k,v,e)}) \,\big\}_{\,v,\; e \in \mathcal{E}_{vis}^{(s,k,v)}}
    = M_{k,\alpha_k}^{(s)}\big(\bm{X}^{(s,k)}\big),
\end{align}
i.e.\ the per-view entity-dropping combined with the per-view ratio $\beta^{(s,k,v)}$ is, in expectation and under the equal-token assumption within each $(k,v)$, equivalent to applying a single mask ratio $\alpha_k$ to modality $k$.

\subsection{LAMAE}
\label{sec:lamae}
During pretraining, we optimize the following objective $\mathcal{L}^{(s)}$ for every hospital stay $s$
\begin{align}
    \mathcal{L}^{(s)} = \sum_{k \in \mathcal{M}_{\text{obs}}^{(s)}} \frac{1}{|\mathcal{V}_{k,\text{obs}}^{(s)}|} \sum_{v \in \mathcal{V}_{k,\text{obs}}^{(s)}} \frac{1}{E_k} \sum_{e \in \mathcal{E}_k} \frac{1}{\alpha_k T_k}
    \sum_{t \notin \mathcal{T}_{k,vis}^{(s)}} \left\| \bm{x}^{(s,k,v,e)}_{t} -  \hat{\bm{x}}^{(s,k,v,e)}_{t} \right\|_2^2,
\end{align}
where $\hat{\bm{x}}^{(s,k,v,e)} = D_{\theta_k}\big(\big[LA_\phi(\bm{Z}^{(s)})\big]^{(k,v,e)}\big)$ and
$\big[LA_\phi(\bm{Z}^{(s)})\big]^{(k,v,e)} = \bm{z}_{out}^{(s,k,v,e)}$ denotes the output embeddings of base entity $e$ in view $v$ of modality $k$.
A key property of our design is that it handles missing modalities by construction, rather than requiring complete cases or imputation as many multimodal methods do. Because the reconstruction objective is applied only over the observed modalities, and the latent-attention (LA) module aggregates a variable-size set of available modality tokens, the model both trains and performs inference on any subset of modalities without architectural change or retraining. Because every output embedding can attend to all observed modalities, each entity-level CLS embedding after the LA module carries information from the whole hospital stay.

For downstream tasks, we represent a hospital stay by the average of the entity-level CLS embeddings after the LA module,
\begin{align}
    \bar{\bm{z}}^{(s)} = \frac{1}{N^{(s)}} \sum_{k \in \mathcal{M}_{\text{obs}}^{(s)}} \sum_{v \in \mathcal{V}_{k,\text{obs}}^{(s)}} \sum_{e \in \mathcal{E}_k} \bm{z}_{\mathrm{cls},out}^{(s,k,v,e)},
\end{align}
where $\bm{z}_{\mathrm{cls},out}^{(s,k,v,e)}$ is the CLS embedding of base entity $e$ in $\bm{Z}_{out}^{(s)}$ and $N^{(s)}$ is the number of observed base entities of stay $s$.

%% file: 04_experiments.tex
\section{Experiments \& Results}
\label{sec:experiments}
Our aim is to evaluate how the LA block influences the pretraining.
We do so by assessing the downstream task performance of various tasks, both on a hospital-stay level (see \Cref{sec:exp_mm}), where we expect multimodal models to benefit, and on unimodal tasks, where unimodal models should have all the required information to solve the respective unimodal tasks (see \Cref{sec:exp_um}).
These two settings reflect the different layers of tasks within the healthcare setting, which range from low-level modality-specific tasks such as left ventricular ejection fraction estimation (LVEF) to tasks that reflect the longitudinal nature of a patient journey within a hospital stay such as in-hospital mortality or diagnosis codes.
We denote by LAMAE$_{\text{MM}}$ the model pretrained on all three modalities, and by LAMAE$_{k}$ and Independent MAE$_{k}$, $k \in \{\text{CXR}, \text{ECG}, \text{Echo}\}$, models pretrained on a single modality $k$.

We study four complementary questions:




\noindent
(i) \textit{Does cross-modal interaction during pretraining improve hospital-stay representations?} \\
LAMAE$_{\text{MM}}$ outperforms Independent MAE$_{\text{MM}}$ on all six hospital-stay tasks and both state-of-the-art baselines on five of six tasks (\Cref{tab:multimodal_results}).

\noindent
(ii) \textit{Does multimodal pretraining outperform unimodal pretraining on hospital-stay tasks?} \\
LAMAE$_{\text{MM}}$ outperforms every unimodal LAMAE$_k$ on the complete test set, demonstrating its advantage in clinically realistic settings with heterogeneous modality availability (\Cref{tab:modality_restriction}). Importantly, LAMAE$_{\text{MM}}$ also outperforms most of the unimodal models on a modality constrained test set across modality configurations, indicating that its gains extend beyond increased coverage and reflect the benefit of the multimodal architecture itself (\Cref{tab:unimodal}).

\noindent
(iii) \textit{Does the LA module improve representations within a single modality?} \\
LAMAE$_{k}$ outperforms Independent MAE$_{k}$ on CXR, Echo, and most ECG targets (\Cref{tab:cxr_macro_auroc,tab:unimodal_echo,tab:unimodal_ecg}).

\noindent
(iv) \textit{Does multimodal pretraining come at the expense of unimodal performance?} \\
No. Finetuned on a single modality, LAMAE$_{\text{MM}}$ matches or exceeds LAMAE$_{k}$ on most unimodal tasks (\Cref{tab:cxr_macro_auroc,tab:unimodal_echo,tab:unimodal_ecg}).

\subsection{Data}
\label{sec:data}
We use the MIMIC-IV database \citep{johnson2023mimic,goldberger2000physiobank}, together with its modality-specific subsets MIMIC-CXR \citep{johnson2019mimic}, MIMIC-IV-ECG \citep{gow2023mimic}, and MIMIC-IV-Echo \citep{gow_mimic-iv-echo_2026} to generate a dataset of multimodal hospital stays.
A hospital stay is a collection of ordered modalities describing the health status of a patient.
Note that not every hospital stay $\mathcal{S}^{(s)}$ is comprised of the same modalities as every patient trajectory potentially requires a different set of modalities to be acquired.
In total, our dataset consists of 658{,}847 unique hospital stays with a total of 227{,}835 CXR studies and 377{,}110 images, 7{,}243 Echo studies and 525{,}328 videos, and 800{,}035 12-lead ECG measurements.

\subsection{Multimodal Evaluation}
\label{sec:exp_mm}
Clinical questions at the level of a hospital stay, such as diagnosis, length of stay or mortality risk, depend on integrating imaging, waveform and clinical information collected during the stay rather than on any single measurement.
We therefore evaluate LAMAE on hospital-stay-level tasks, where this kind of cross-modal reasoning is most directly required.

\textit{Setup.} 
We evaluate on six hospital-stay-level tasks from four task groups: in-hospital mortality, ICD-10 Chapter~IX diagnosis coding \citep{world_health_organization_icd-10_2004}, DRG classification (code, severity and mortality risk) \citep{fetter_case_1980}, and length of stay.
We compare against two state-of-the-art baselines, ProbMED \citep{gao_probmed_2025} and MedSigLip \citep{sellergren_medgemma_2026} (details in \Cref{app:experiments}).
In addition, we include an Independent MAE that is trained on all three modalities with the same data and training budget as LAMAE, but without the LA module and the entity masking (\Cref{tab:app_pretraining});
its modalities therefore never interact during pretraining.
To test whether multimodal pretraining and multimodal input are beneficial, we further pretrain and finetune one LAMAE per modality (LAMAE$_{\text{CXR}}$, LAMAE$_{\text{ECG}}$, LAMAE$_{\text{Echo}}$).
Each of these unimodal models can score only the stays that contain its modality; all other test stays receive the training-set prior as prediction, so that every model is evaluated on the same test set (\Cref{app:experiments}).
All models are finetuned on 50{,}000 labeled hospital stays using the same finetuning configuration (\Cref{tab:app_finetuning_multimodal}). We select the checkpoint with the lowest validation loss and report mean and standard deviation over three seeds on the test set.

\textit{Results.}
\Cref{tab:multimodal_results} shows that LAMAE$_{\text{MM}}$ achieves the best AUROC on five of the six tasks, outperforming ProbMED, MedSigLip and the Independent MAE on in-hospital mortality, ICD-10, DRG severity, DRG mortality risk and length of stay.
On DRG code prediction, the two baselines are slightly ahead (59.39 and 59.38 vs.\ 58.92), with differences of about one standard deviation.
The comparison with the Independent MAE highlights the contribution of cross-modal interaction during pretraining: LAMAE improves on every task, by 0.8 to 1.5 percentage points.
The Independent MAE on its own is already on par with the state-of-the-art baselines, ahead on mortality and length of stay and behind on ICD-10 and DRG code, which suggests that LAMAE's advantage stems from modeling cross-modal interactions rather than from the multimodal pretraining data alone.
\Cref{tab:modality_restriction} compares LAMAE$_{\text{MM}}$ with LAMAE models pretrained and finetuned on a single modality.
LAMAE$_{\text{MM}}$ outperforms all unimodal models on every task.
The strongest unimodal model, LAMAE$_{\text{ECG}}$, is based on the most widely available modality, but is still behind LAMAE$_{\text{MM}}$ by 2.7 to 7.4 percentage points, with the largest gaps on ICD-10 and length of stay.
LAMAE$_{\text{CXR}}$ and especially LAMAE$_{\text{Echo}}$ remain close to chance, because they can score only the small fraction of stays that contain their modality.
These gaps therefore combine two effects: the information contributed by the other modalities, and the ability of LAMAE$_{\text{MM}}$ to make a prediction for any stay with at least one observed modality.
Removing cross-modal interaction (Independent MAE) and removing modalities (unimodal LAMAEs) both degrade performance, showing that multimodal pretraining with the LA module and access to multiple modalities are both needed for the best hospital-stay performance.

\begin{table}[t]
\caption{\textbf{Hospital Stay Tasks.} We finetune all methods using 50{,}000 labeled samples. Every sample has at least one ECG, Echo, or CXR measurement.
At test, the model receives the set of modalities that is available for each hospital stay.
We report the AUROC (\%) or macro average of AUROC where applicable (mean $\pm$ std).}
\label{tab:multimodal_results}
\begin{center}
\resizebox{\textwidth}{!}{%
\begin{tabular}{lcccccc}
\toprule
&\multicolumn{1}{c}{\raisebox{-0.1em}{\includegraphics[height=1em]{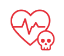}} Mortality}  &\multicolumn{1}{c}{\raisebox{-0.1em}{\includegraphics[height=1em]{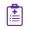}}ICD10}  &\multicolumn{3}{c}{\raisebox{-0.1em}{\includegraphics[height=1em]{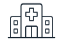}}DRG} &\multicolumn{1}{c}{\raisebox{-0.15em}{\includegraphics[height=1em]{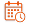}}Length of Stay}
\\
\cmidrule(lr){2-2} \cmidrule(lr){3-3} \cmidrule(lr){4-6} \cmidrule(lr){7-7}
\multicolumn{1}{c}{Model} & {\small \textit{In-hospital}} & {\small \textit{Ch. IX}}& {\small \textit{Code}} & {\small \textit{Severity}} & {\small \textit{Mortality Risk}} & {\small \textit{In-hospital}}\\
\midrule
MedSigLip   & $89.62 {\scriptstyle \pm 0.38}$ & $82.06 {\scriptstyle \pm 0.18}$ & $\mathbf{59.39} {\scriptstyle \pm 0.38}$ & $68.75 {\scriptstyle \pm 0.22}$ & $70.65 {\scriptstyle \pm 0.06}$ & $79.20 {\scriptstyle \pm 0.44}$ \\
ProbMED     & $90.20 {\scriptstyle \pm 0.66}$ & $82.56 {\scriptstyle \pm 0.08}$ & $59.38 {\scriptstyle \pm 0.35}$ & $69.31 {\scriptstyle \pm 0.28}$ & $71.36 {\scriptstyle \pm 0.34}$ & $79.81 {\scriptstyle \pm 0.38}$ \\
Independent MAE$_{\text{MM}}$ & $90.51 {\scriptstyle \pm 0.47}$ & $81.66 {\scriptstyle \pm 0.11}$ & $57.46 {\scriptstyle \pm 0.23}$ & $69.08 {\scriptstyle \pm 0.36} $ & $71.14 {\scriptstyle \pm 0.23}$ & $80.87 {\scriptstyle \pm 0.25}$ \\
\midrule
LAMAE$_{\text{MM}}$ & $\mathbf{91.59} {\scriptstyle \pm 0.37}$ & $\mathbf{83.02} {\scriptstyle \pm 0.18}$ & $58.92 {\scriptstyle \pm 0.31}$ & $\mathbf{70.29} {\scriptstyle \pm 0.24} $ & $\mathbf{72.21} {\scriptstyle \pm 0.43} $ & $\mathbf{81.70} {\scriptstyle \pm 0.26}$ \\
\bottomrule





\end{tabular}
}
\end{center}
\end{table}

\begin{table}[t]
\caption{\textbf{Unimodal vs. Multimodal Models.} The subscript gives the set of modalities the model was pretrained on and takes as input. Test samples that lack an accepted modality are assigned the training-set prior, so every row is scored on the same test set.
We report AUROC (\%), or the macro average of AUROC where applicable (mean $\pm$ std).}
\label{tab:modality_restriction}
\begin{center}
\resizebox{\textwidth}{!}{%
\begin{tabular}{lcccccc}
\toprule
&\multicolumn{1}{c}{\raisebox{-0.1em}{\includegraphics[height=1em]{figures/icon_mortality.png}} Mortality}  &\multicolumn{1}{c}{\raisebox{-0.1em}{\includegraphics[height=1em]{figures/icon_icd10.png}}ICD10}  &\multicolumn{3}{c}{\raisebox{-0.15em}{\includegraphics[height=1em]{figures/icon_drg.png}}DRG} &\multicolumn{1}{c}{\raisebox{-0.1em}{\includegraphics[height=1em]{figures/icon_los.png}}Length of Stay}
\\
\cmidrule(lr){2-2} \cmidrule(lr){3-3} \cmidrule(lr){4-6} \cmidrule(lr){7-7}
\multicolumn{1}{c}{Model} & {\small \textit{In-hospital}} & {\small \textit{Ch. IX}}& {\small \textit{Code}} & {\small \textit{Severity}} & {\small \textit{Mortality Risk}} & {\small \textit{In-hospital}} \\
\midrule
LAMAE$_{\text{Echo}}$ & $49.80 {\scriptstyle \pm 0.24}$ & $50.45 {\scriptstyle \pm 0.09}$ & $42.36 {\scriptstyle \pm 0.05}$ & $50.20 {\scriptstyle \pm 0.14}$ & $50.31 {\scriptstyle \pm 0.09}$ & $50.47 {\scriptstyle \pm 0.03}$ \\
LAMAE$_{\text{CXR}}$ & $61.03 {\scriptstyle \pm 0.20}$ & $56.53 {\scriptstyle \pm 0.05}$ & $51.25 {\scriptstyle \pm 0.11}$ & $52.91 {\scriptstyle \pm 0.16}$ & $53.35 {\scriptstyle \pm 0.06}$& $57.48 {\scriptstyle \pm 0.04}$ \\
LAMAE$_{\text{ECG}}$ & $86.22 {\scriptstyle \pm 0.69}$ & $75.67 {\scriptstyle \pm 0.52}$ & $55.65 {\scriptstyle \pm 0.68}$ & $67.50 {\scriptstyle \pm 0.11}$ & $69.56 {\scriptstyle \pm 0.08}$ & $74.47 {\scriptstyle \pm 0.21}$ \\

\midrule
LAMAE$_{\text{MM}}$ & $\mathbf{91.59} {\scriptstyle \pm 0.37}$ & $\mathbf{83.02} {\scriptstyle \pm 0.18}$ & $\mathbf{58.92} {\scriptstyle \pm 0.31}$ & $\mathbf{70.29} {\scriptstyle \pm 0.24} $ & $\mathbf{72.21} {\scriptstyle \pm 0.43} $ & $\mathbf{81.70} {\scriptstyle \pm 0.26}$\\

\bottomrule
\end{tabular}
}
\end{center}
\end{table}


\subsection{Unimodal Evaluation}
\label{sec:exp_um}
While we propose LAMAE mainly for its multimodal capabilities, its hierarchical approach can be leveraged equally well for unimodal problems and datasets where some structure is present to be leveraged.
Therefore, we evaluate LAMAE on unimodal tasks too, validating the unimodal performance of the proposed LAMAE approach. Additionally, we report ICD-10 classification performance, but unlike in \Cref{tab:modality_restriction}, we evaluate only on records where the examined modality is present, isolating how well the model uses that modality.
For all modalities, we select the epoch with the lowest validation loss and report test performance over three seeds.

\begin{table}[ht]
\centering
\caption{\textbf{Unimodal Results.} We evaluate the capabilities of the latent attention module to extract meaningful representations also for unimodal data.
We compare a LAMAE that is trained only on a single data type to an independent MAE that is not leveraging any structure in the data.
We report finetuning results on the test set (mean $\pm$ std over 3 seeds).
We compare the models on relevant unimodal tasks and the classification of ICD10 codes (see \Cref{tab:multimodal_results}).}
\label{tab:unimodal}
\begin{subtable}[t]{0.48\textwidth}
    \centering
    \caption{\textbf{CXR.} We use the 14 CheXpert labels \citep{irvin_chexpert_2019} to assess the understanding of CXR data and report the macro AUROC (\%, $\uparrow$).}
    \label{tab:cxr_macro_auroc}
    \resizebox{\linewidth}{!}{%
    \begin{tabular}{lcc}
    \toprule
    \textbf{Method} & \multicolumn{1}{c}{CXR Tasks} & \multicolumn{1}{c}{\raisebox{-0.1em}{\includegraphics[height=1em]{figures/icon_icd10.png}}ICD10} \\
    \midrule
    Independent MAE$_{\text{CXR}}$ & ${82.58}  {\scriptstyle \pm 0.07}$ & ${85.03}  {\scriptstyle \pm 0.05}$ \\
    LAMAE$_{\text{CXR}}$ & $ {83.29}  {\scriptstyle \pm 0.04}$ & ${85.56}  {\scriptstyle \pm 0.13}$ \\
    LAMAE$_{\text{MM}}$ & $\mathbf{83.42}  {\scriptstyle \pm 0.05}$ & $\mathbf{85.88}  {\scriptstyle \pm 0.13}$\\ 
    \bottomrule
    \end{tabular}
    }
\end{subtable}
\hfill
\begin{subtable}[t]{0.48\textwidth}
    \centering
    \caption{
    \textbf{Echo.} We evaluate the Echo-specific performance in the left ventricular ejection fraction (LVEF) regression task, where we report the $R^2$ score ($\uparrow$).
    }
    \label{tab:unimodal_echo}
    \resizebox{\linewidth}{!}{%
    \begin{tabular}{lcc}
    \toprule
    \textbf{Method} & \multicolumn{1}{c}{LVEF } & \multicolumn{1}{c}{\raisebox{-0.1em}{\includegraphics[height=1em]{figures/icon_icd10.png}}ICD10}\\
    \midrule


    Independent MAE$_{\text{Echo}}$          & $0.273 {\scriptstyle \pm 0.008}$ & $57.49 {\scriptstyle \pm 0.40}$\\
    LAMAE$_{\text{Echo}}$                    & $\mathbf{0.414} {\scriptstyle \pm 0.055}$ & $\mathbf{59.21} {\scriptstyle \pm 0.71}$\\
    LAMAE$_{\text{MM}}$            & $\mathbf{0.414} {\scriptstyle \pm 0.009}$ & $59.10 {\scriptstyle \pm 0.40}$\\

    \bottomrule
    \end{tabular}
    }
\end{subtable}
\\[1.5ex]
\begin{subtable}[t]{\textwidth}
    \centering
    \caption{\textbf{ECG.}
    We evaluate the ECG interval regression, reported in $R^2$ score ($\uparrow$), alongside the ECG-specific ICD-10 task AUROC (\%) ($\uparrow$).
    }
    \label{tab:unimodal_ecg}
    \resizebox{\linewidth}{!}{
    \begin{tabular}{lccccccc}
    \toprule
    \textbf{Method} & RR & PR & P wave & QRS & QT & QTc & \multicolumn{1}{c}{\raisebox{-0.1em}{\includegraphics[height=1em]{figures/icon_icd10.png}}ICD10}\\
    \midrule

    Independent MAE$_{\text{ECG}}$ & $0.91{\scriptstyle \pm 0.00}$ & $0.13{\scriptstyle \pm 0.05}$ & $-0.36{\scriptstyle \pm 0.28}$ & $0.51{\scriptstyle \pm 0.17}$ & $0.58{\scriptstyle \pm 0.07}$ & $0.23{\scriptstyle \pm 0.05}$ & $70.08{\scriptstyle \pm 1.71}$ \\
    \addlinespace
    LAMAE$_{\text{ECG}}$ & $0.95{\scriptstyle \pm 0.01}$ & $0.43{\scriptstyle \pm 0.02}$ & $-0.03{\scriptstyle \pm 0.06}$ & $\mathbf{0.53}{\scriptstyle \pm 0.33}$ & $0.81{\scriptstyle \pm 0.02}$ & $\mathbf{0.54}{\scriptstyle \pm 0.05}$ & $76.87{\scriptstyle \pm 0.52}$ \\
    LAMAE$_{\text{MM}}$ & $\mathbf{0.98}{\scriptstyle \pm 0.01}$ & $\mathbf{0.67}{\scriptstyle \pm 0.03}$ & $\mathbf{-0.02}{\scriptstyle \pm 0.08}$ & $0.51{\scriptstyle \pm 0.10}$ & $\mathbf{0.85}{\scriptstyle \pm 0.05}$ & $0.41{\scriptstyle \pm 0.32}$ & $\mathbf{79.13}{\scriptstyle \pm 0.30}$ \\
    \bottomrule
    \end{tabular}
    }
\end{subtable}
\end{table}
\noindent
\textbf{Chest X-Ray}

\textit{Setup.} We evaluate on the 14-label CheXpert finding classification task described above, with each CXR study capped at a maximum of three views. Alongside the view-based Independent MAE$_{\text{CXR}}$ baseline, trained solely on CXR, we evaluate LAMAE$_{\text{CXR}}$, and additionally report LAMAE$_{\text{MM}}$, the same fully multimodal-pretrained checkpoint used in the Echo evaluation, finetuned and tested with its input restricted to CXR only. Pretraining uses all available CXR studies for 1000 epochs; finetuning runs for 50 epochs.

\textit{Result.}
Table~\ref{tab:cxr_macro_auroc} reports the results. LAMAE$_{\mathrm{CXR}}$ improves over the Independent MAE$_{\mathrm{CXR}}$ on both CXR findings and ICD-10 classification, increasing macro AUROC from 82.58 to 83.29 and from 85.03 to 85.56, respectively. LAMAE$_{\mathrm{MM}}$ further improves performance to 83.42 and 85.88, despite being evaluated with CXR input only. This indicates that multimodal pretraining improves the CXR representation even when no additional modalities are available at finetuning or test time.
\Cref{fig:cxr_radar} shows how this aggregate score breaks down across the 14 individual CheXpert findings.

\noindent
\textbf{Echocardiogram}

\textit{Setup.} We predict left ventricular ejection fraction (LVEF), taken from the structured measurements accompanying the MIMIC-IV-Echo studies \citep{gow_mimic-iv-echo_2026}. Alongside the frame-based Independent MAE$_{\text{Echo}}$ baseline, we evaluate LAMAE$_{\text{Echo}}$ and additionally report LAMAE$_{\text{MM}}$, the same multimodal-pretrained checkpoint used in the evaluation of \cref{sec:exp_mm}, finetuned and tested with its input restricted to Echo only. From each video (selection described in \Cref{app:experiments}), we sample 8 frames with a stride of 4 at a random offset. Hyperparameters are listed in \Cref{tab:app_pretraining,tab:app_finetuning_unimodal}.

\textit{Result.} \Cref{tab:unimodal_echo} reports the results. Both LAMAE variants substantially improve over the baseline, raising LVEF $R^2$ from $0.273$ to $0.414$ and ICD-10 AUROC from $57.49$ to $59.21$ and $59.10$, with the two variants within seed variance of each other.


\textbf{Electrocardiogram}

\textit{Setup.} We derive six global interval measurements from the MIMIC-IV-ECG machine annotations \citep{strodthoff2024mimic}: the RR interval, taken directly from the annotations, and the PR interval, P-wave duration, QRS duration and QT interval, obtained as differences of the annotated wave onsets and offsets. 
PR is the interval from P onset to QRS onset, P wave from P onset to P end, QRS from QRS onset to QRS end, and QT from QRS onset to T end.
The rate-corrected QTc interval is computed from QT and RR using Bazett's formula, following \citet{Plagwitz2025QTcNetDeepLearning}. 
Alongside the lead-based Independent MAE$_{\text{ECG}}$ baseline, we evaluate LAMAE$_{\text{ECG}}$ and additionally report LAMAE$_{\text{MM}}$, the same multimodal-pretrained checkpoint used in the evaluation of \cref{sec:exp_mm}, finetuned and tested with its input restricted to ECG only.
Hyperparameters are listed in \Cref{tab:app_pretraining,tab:app_finetuning_unimodal}.

\textit{Result.} \Cref{tab:unimodal_ecg} and \Cref{app:experiments} report the results.
Both LAMAE variants substantially improve over the baseline on PR, QT and QTc, and on ICD-10 classification ($70.08$ to $76.87$ and $79.13$ AUROC), with LAMAE$_{\text{MM}}$ performing best on most targets. QRS duration is within seed variance across all models, and P-wave duration is not recovered by any model ($R^2 \le 0$).

\noindent
\textbf{Summary.}
Overall, LAMAE improves representation learning even in the unimodal setting, showing that its benefits extend beyond multimodal integration through its ability to naturally leverage the inherent structure of medical data.

%% file: 05_conclusion.tex
\section{Conclusion}
\label{sec:conclusion}
We introduced Latent-Attention Masked Autoencoders (LAMAE), a multimodal, structure-aware masked autoencoder that learns patient-level representations from ECG, Echo, and CXR during self-supervised pretraining.
Instead of fusing modalities only at the prediction stage, LAMAE exchanges information across modalities in the latent space already during pretraining.
It does so through a shared latent-attention (LA) module operating over a study--view--entity hierarchy, naturally handling heterogeneous data, variable numbers of observations, and missing modalities within a single reconstruction objective.
Pretrained on more than $500{,}000$ MIMIC-IV hospital stays, LAMAE outperforms an otherwise identical model without cross-modal interaction, models pretrained on a single modality, and strong contrastive and vision--language baselines on hospital-stay-level tasks.
The benefits of modeling structure extend to the unimodal setting: the LA module improves over independent MAEs on CXR, Echo, and ECG tasks, and the multimodal-pretrained model matches or exceeds its modality-specific counterparts when finetuned on a single modality, indicating that cross-modal pretraining does not come at the expense of unimodal performance.
This work has limitations: all experiments use data from a single institution, Echo is available for only a small fraction of hospital stays, and some fine-grained targets, such as P-wave duration, are not recovered by any model.
Promising next steps include validating cross-institutional transfer and extending the LA framework to further modalities such as laboratory time series and clinical notes.
By learning representations that reflect the multimodal, structured nature of clinical reasoning, LAMAE offers a step toward foundation models that support the holistic, integrative patient assessment clinicians perform in practice.

%% file: 06_appendix.tex

\section{Implementation Details}
\label{app:implementation}

\paragraph{Instantiating the study--view--entity hierarchy.}
Each sample $\mathcal{S}^{(s)}$ is one MIMIC-IV hospital stay, with modalities $\mathcal{M}=\{\text{CXR},\text{ECG},\text{Echo}\}$. Hospital stays contain only a subset of these modalities, as depicted in Figure \ref{fig:modality_upset}. Table~\ref{tab:app_hierarchy} shows how each modality is mapped onto the hierarchy of hospital stays as outlined in \Cref{sec:method}.
For \textbf{CXR}, each radiograph of the stay is a view (frontal, lateral or other). The image itself is the base entity, so $E_{\text{CXR}}=1$. We use up to three views per study, sampled at random if more are available.
For \textbf{ECG}, a study is one 12-lead recording sampled at random from the stay. Its base entities are the 12 leads, each a 10\,s signal at 500\,Hz.
For \textbf{Echo}, a study is one ultrasound video sampled at random from the stay's examination. Its base entities are 8 frames, linearly spaced over a random window of up to 32 consecutive frames.
Missing modalities and views are padded and ignored in both attention and the loss.

\begin{table}[h]
\centering
\caption{\textbf{Study--view--entity hierarchy for each modality.} $V_k$: maximum number of views per study; $E_k$: base entities per view; $T_k$: tokens per base entity.}
\label{tab:app_hierarchy}
\small
\begin{tabular}{lccc}
\toprule
 & CXR & ECG & Echo \\
\midrule
View $X^{(s,k,v)}$ & radiograph & 12-lead recording & video \\
Base entity $\boldsymbol{x}^{(s,k,v,e)}$ & image & lead & frame \\
$V_k$ / $E_k$ & 3 / 1 & 1 / 12 & 1 / 8 \\
Input per entity & $1\times224\times224$ & $1\times5000$ & $3\times224\times224$ \\
Patch size & $16\times16$ & 50 samples & $14\times14$ \\
$T_k$ & 196 & 100 & 256 \\
\bottomrule
\end{tabular}
\end{table}

\paragraph{Architecture.}
Each modality has its own encoder $E_{\phi_k}$ and decoder $D_{\theta_k}$. The encoders follow ViT-B (12 blocks, width 768, 12 heads), and their transformer blocks are initialized from DINOv3~\citep{simeoni2025dinov3}. The decoders are lighter, with 4 blocks of the same width. Patches are embedded by a linear layer, and each base entity gets its own CLS token. We add learned embeddings for patch position, entity (ECG lead, echo frame) and view (CXR projection, echo view).
The implementation of the latent attention module $LA_\phi$ is a single layer transformer block that operates on the concatenated patch and CLS tokens of all observed modalities.
It additionally carries a learned multimodal CLS token, which is not used for downstream tasks. Learned modality embeddings let it distinguish tokens from different modalities. The Independent MAE uses the same encoders and decoders but no LA module.
\begin{wrapfigure}{r}{0.4\textwidth}
    \centering
    \includegraphics[width=0.4\textwidth]{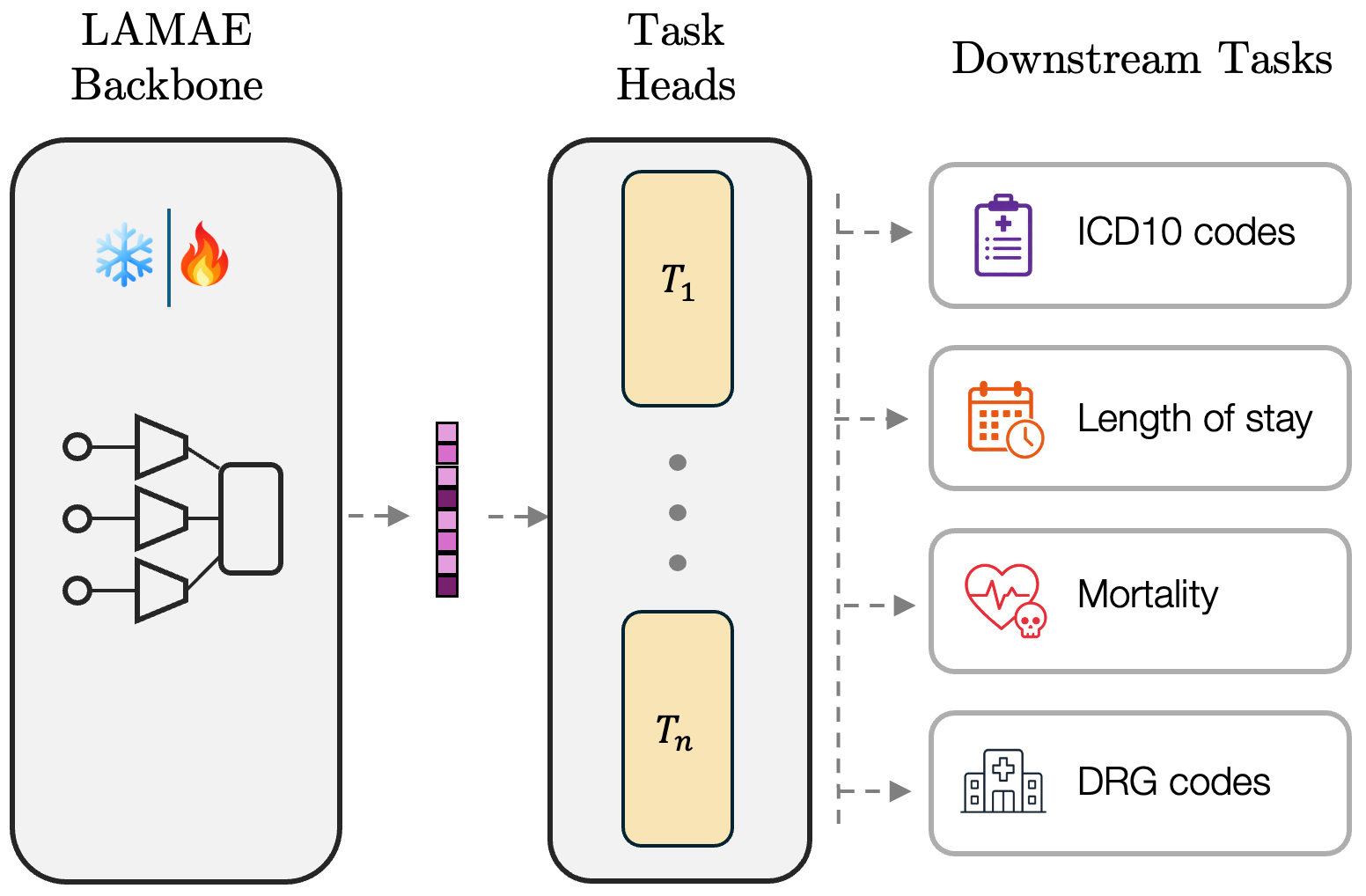}
    \caption{\textbf{Finetuning pipeline.} The hospital-stay downstream tasks are shown as an illustrative example. The same finetuning procedure is used for the unimodal downstream tasks.}
    \label{fig:finetuning}
\end{wrapfigure}
\paragraph{Masking and objective.}
We mask a fraction $\alpha_k$ of the tokens of each view. For LAMAE, part of this masking consists of dropping complete entities. The remaining visible entities are masked more lightly, so that the overall ratio stays at $\alpha_k$. Table~\ref{tab:app_pretraining} lists $\alpha_k$ and the number of dropped entities for each model. The loss is the mean squared error over masked patches, averaged within each modality and then summed over the observed modalities.

\paragraph{Downstream heads.}
For finetuning, we discard the decoders and disable masking. A stay is represented by the average of the entity-level CLS embeddings after the LA module (\cref{sec:lamae}); for the Independent MAE, we average the entity-level CLS embeddings of the encoders. In preliminary experiments, this average performed slightly better than the multimodal CLS token of the LA module, consistent with \citep{he2022masked}, who find that average pooling works similarly well as a class token. A linear head follows (preceded by LayerNorm). The losses are binary cross-entropy for binary and multi-label tasks, cross-entropy for multi-class tasks and the L1 loss for regression tasks. For unimodal evaluation of LAMAE\textsubscript{MM}, we feed only the respective modality.

\paragraph{Preprocessing and augmentation.}
Images are normalized with dataset statistics. CXR training uses random resized crops, horizontal flips and brightness/contrast jitter. Echo training uses random resized crops and brightness jitter, without flips. ECGs are z-score normalized and augmented with STAR time warping ~\citep{nemati2025comparative}.

\section{Training Details}
\label{app:training}

All models are trained with AdamW (weight decay 0.05), gradient clipping at 1.0 and FP32 precision. The learning rate is scaled with the global batch size $B$ as $\eta=\eta_{\text{base}}\sqrt{B}/256$. Pretraining uses a linear warm-up, a constant learning rate and a short final cooldown. Finetuning uses a linear warm-up followed by cosine decay.~\Cref{tab:app_pretraining} summarizes the main hyperparameters for pretraining. We use the final pretraining checkpoint. For finetuning, we select the checkpoint with the lowest validation loss and report mean and standard deviation over three seeds. All experiments ran on NVIDIA GH200 GPUs (four per node). Pretraining LAMAE with all the available modalities required approximately \textit{4198} GPU hours, while full finetuning for a single downstream task and seed required approximately \textit{4.33} GPU hours.

\begin{table}[h]
\centering
\caption{\textbf{Pretraining hyperparameters.} LA: LAMAE; Ind.: Independent MAE. For the hospital-stay comparison (\Cref{tab:modality_restriction}), LAMAE$_{\text{CXR}}$ is pretrained for 3000 epochs to give the unimodal CXR baseline its strongest configuration; all models in the unimodal evaluation (\Cref{tab:unimodal}) use 1000-epoch checkpoints for fairness.}
\label{tab:app_pretraining}
\small
\setlength{\tabcolsep}{4pt}
\begin{tabular}{lcccccccc}
\toprule
 & \multicolumn{2}{c}{Multimodal} & \multicolumn{2}{c}{CXR} & \multicolumn{2}{c}{Echo} & \multicolumn{2}{c}{ECG} \\
\cmidrule(lr){2-3}\cmidrule(lr){4-5}\cmidrule(lr){6-7}\cmidrule(lr){8-9}
 & LA & Ind. & LA & Ind. & LA & Ind. & LA & Ind. \\
\midrule
Epochs            & 1000 & 1000 & 1000 & 3000/1000 & 1000 & 1000 & 1000 & 1000 \\
GPUs              & 128 & 128 & 128 & 128 & 16 & 16 & 128 & 128 \\
Global batch size & 4096 & 4096 & 4096 & 4096 & 256 & 256 & 8192 & 8192 \\
Base LR           & 5e-4 & 5e-4 & 1e-3 & 1e-3 & 4e-3 & 4e-3 & 1e-4 & 1e-4 \\
Mask ratio $\alpha_k$ & 0.95 & 0.90 & 0.90 & 0.90 & 0.875 & 0.875 & 0.90 & 0.90 \\
Dropped ECG leads   & 8--11 & -- & -- & -- & -- & -- & 8--11 & -- \\
Dropped Echo frames & 6--7 & -- & -- & -- & 3--4 & -- & -- & -- \\
\bottomrule
\end{tabular}
\end{table}

\begin{table}[h]
\centering
\caption{\textbf{Finetuning hyperparameters for unimodal downstream tasks.} 
All settings start from the final pretraining checkpoint, fully finetune the model with a linear warm-up followed by cosine decay, and are repeated with three seeds.}
\label{tab:app_finetuning_unimodal}
\small
\setlength{\tabcolsep}{3pt}
\begin{tabular}{lcc|cc|cc}
\toprule
 & CXR tasks & CXR ICD-10 
 & Echo LVEF & Echo ICD-10 
 & ECG intervals & ECG ICD-10 \\
\midrule
Training samples  & all   & 50k  & 50k   & 50k  & 50k  & 50k \\
Epochs            & 50    & 50  & 500   & 40  & 20   & 20 \\
GPUs              & 32    & 16  & 32    & 32  & 128  & 128 \\
Global batch size & 1024  & 512  & 256   & 256  & 4096 & 4096 \\
Base LR           & 1e-4  & 1e-4  & 5e-4  & 5e-4  & 1e-4 & 1e-4 \\
Warm-up steps     & 500   & 500  & 300   & 300   & 300  & 300 \\
Loss              & BCE   & BCE & L1    & BCE & L1   & BCE \\
\bottomrule
\end{tabular}
\end{table}

\begin{table}[h]
\centering
\caption{\textbf{Finetuning hyperparameters for hospital-stay downstream tasks.}
The same configuration is used for all six tasks: in-hospital mortality, ICD-10 Chapter~IX, DRG code, DRG severity, DRG mortality risk, and length of stay. All experiments use 50{,}000 labeled hospital stays, start from the final pretraining checkpoint, and are repeated over three seeds.}
\label{tab:app_finetuning_multimodal}
\small
\begin{tabular}{lc}
\toprule
 & Hospital-stay tasks \\
\midrule
Training samples  & 50k \\
Epochs            & 10 \\
GPUs              & 32 \\
Global batch size & 256 \\
Base LR           & 5e-4 \\
Warm-up steps     & 500 \\
\bottomrule
\end{tabular}
\end{table}

\section{Data}
\label{app:data_distributions}
\paragraph{Dataset splitting.} We split the data at the patient level into training, validation and test sets (80/10/10), so that all stays of a given patient fall into the same split and no patient appears in more than one. Throughout this work, we use the term “hospital stay” to collectively refer to MIMIC emergency department, inpatient, and outpatient encounters. Of the 1{,}267{,}017 stays in the MIMIC cohort, 608{,}170 (48\%) contain no ECG, echocardiogram or chest X-ray and are excluded from the analysis. The remaining stays give 138{,}250 patients (527{,}805 stays) for training, 17{,}263 patients (65{,}296 stays) for validation and 17{,}321 patients (65{,}746 stays) for testing. The same split is used for pretraining and all downstream fine-tuning, so no test patient is seen at any stage of training.

\begin{figure}[h]
    \centering
    \includegraphics[width=0.95\linewidth]{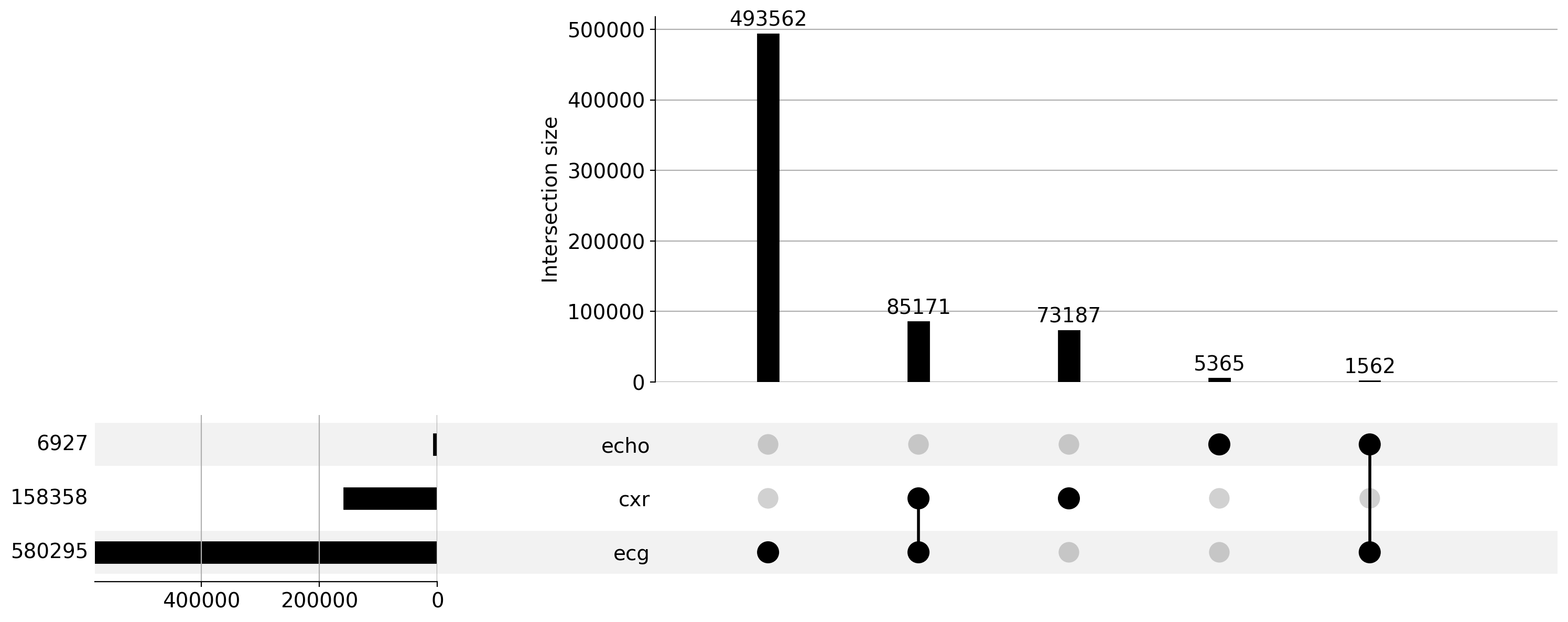}
    \caption{\textbf{Modality availability across hospital stays.}
    UpSet plot showing the availability and overlap of ECG, CXR, and Echo
    across hospital stays in the multimodal MIMIC-IV cohort. Horizontal bars
    report the total number of hospital stays containing each modality, while
    vertical bars report the number of hospital stays for each modality
    combination.}
    \label{fig:modality_upset}
\end{figure}

\paragraph{Modality availability.}
Figure~\ref{fig:modality_upset} summarizes the availability and overlap of
CXR, ECG, and Echo across the hospital-stay cohort.
Modality availability is highly heterogeneous: hospital stays may contain
any subset of the three modalities rather than a complete multimodal
examination. This distribution motivates our formulation of each hospital
stay as a variable-size set of observed modalities and the explicit handling
of missing modalities in LAMAE.

\begin{figure}[t]
    \centering
    \begin{subfigure}[b]{0.75\textwidth}
        \centering
        \includegraphics[width=\textwidth]{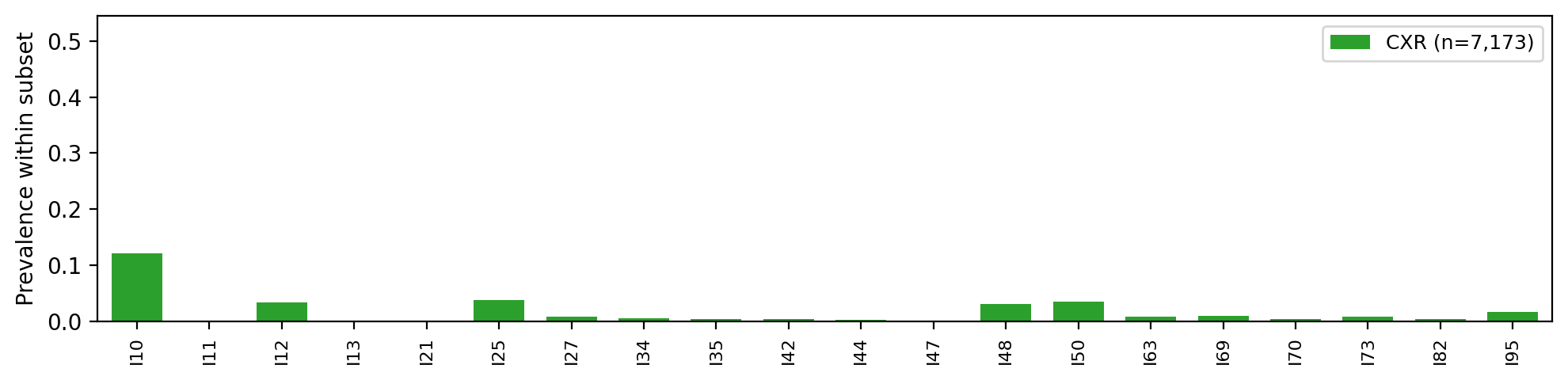}
        \caption{CXR}
        \label{fig:icd10_dist_cxr}
    \end{subfigure}

    \begin{subfigure}[b]{0.75\textwidth}
        \centering
        \includegraphics[width=\textwidth]{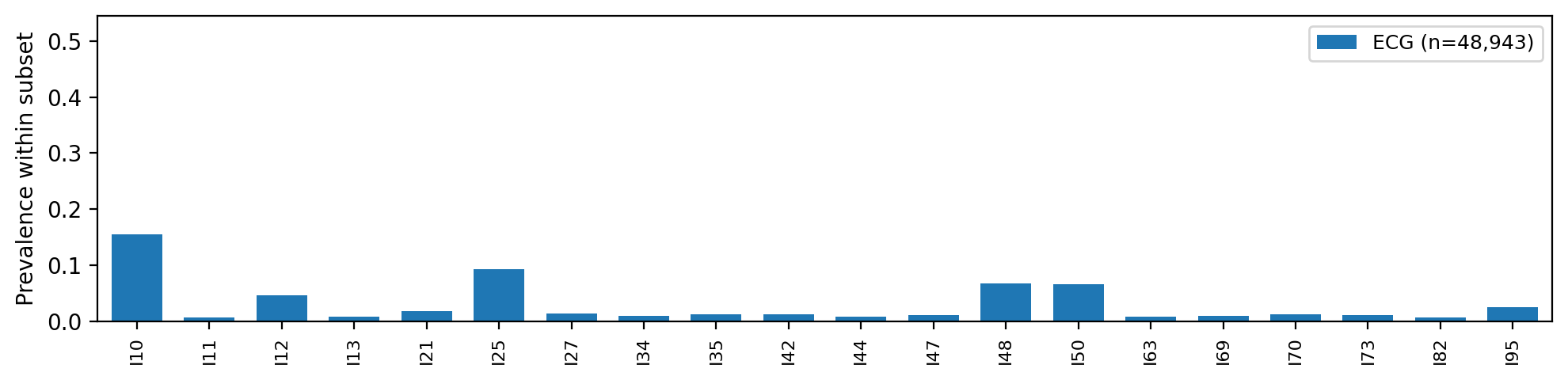}
        \caption{ECG}
        \label{fig:icd10_dist_ecg}
    \end{subfigure}

    \begin{subfigure}[b]{0.75\textwidth}
        \centering
        \includegraphics[width=\textwidth]{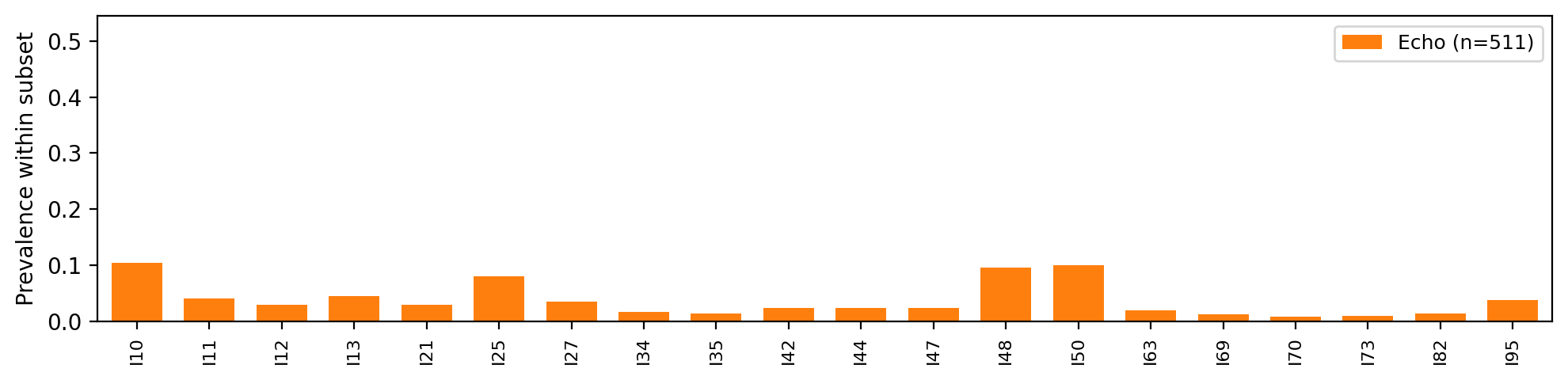}
        \caption{Echo}
        \label{fig:icd10_dist_echo}
    \end{subfigure}

    \begin{subfigure}[b]{0.75\textwidth}
        \centering
        \includegraphics[width=\textwidth]{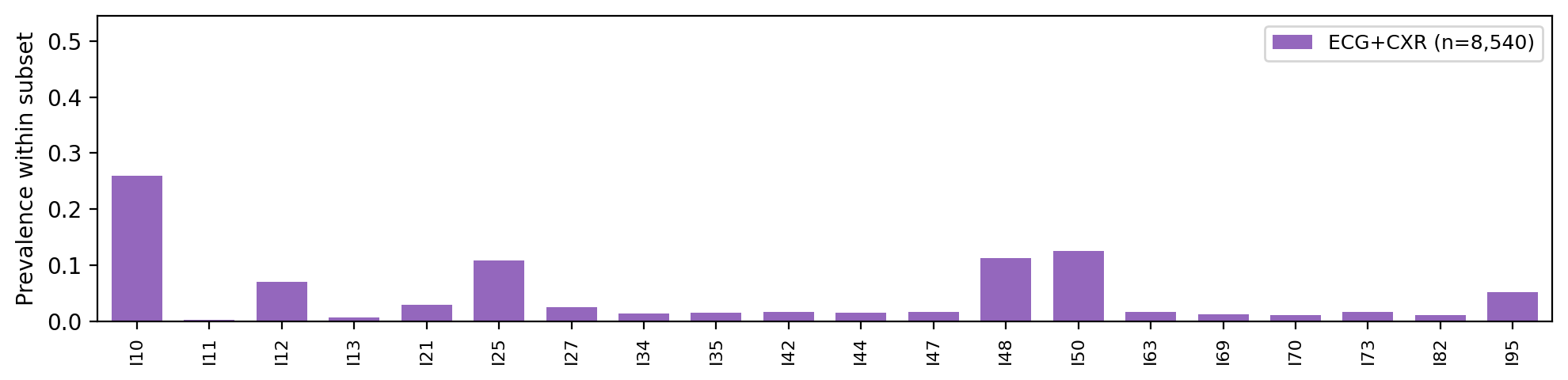}
        \caption{ECG + CXR}
        \label{fig:icd10_dist_ecg_cxr}
    \end{subfigure}

    \begin{subfigure}[b]{0.75\textwidth}
        \centering
        \includegraphics[width=\textwidth]{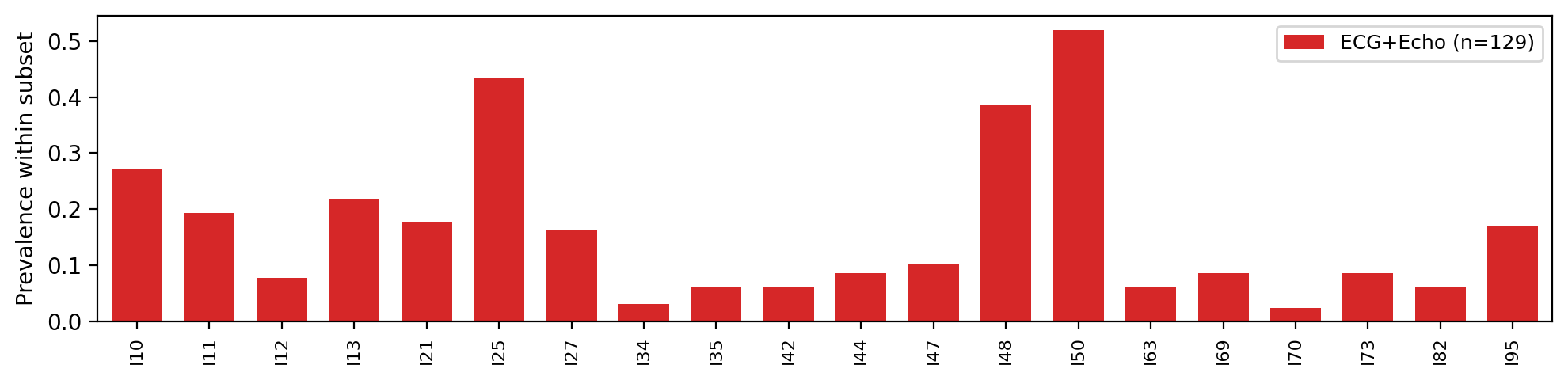}
        \caption{ECG + Echo}
        \label{fig:icd10_dist_ecg_echo}
    \end{subfigure}

    \caption{\textbf{ICD-10 Chapter IX label distributions across modality subsets.}
    Prevalence and number of positive hospital stays for each retained
    ICD-10 Chapter IX diagnosis code or code group, stratified by available
    modality subset. The distributions illustrate both the differences in
    sample availability across modality subsets and the class imbalance of
    the downstream ICD-10 classification task.}
    \label{fig:icd10_modality_distributions}
\end{figure}

\paragraph{ICD-10 label distributions across modality subsets.}
Figure~\ref{fig:icd10_modality_distributions} shows the distribution of
ICD-10 Chapter IX labels across hospital stays with different available
modalities. For each modality subset, we report the prevalence and number of
positive hospital stays for each retained diagnosis code or code group.
Although the absolute number of available samples varies substantially across
modality subsets, the plots provide context on the diagnostic composition of
the corresponding cohorts and on the class imbalance of the ICD-10
classification task. These distributions complement the per-label ICD-10
classification results reported in Figure~\ref{fig:icd10_radar}.

\section{Experiments}
\label{app:experiments}
\paragraph{Baselines}
ProbMED is trained on combinations of CXR, ECG, Echo, and corresponding text samples using a contrastive training objective; it is also trained on data coming from the MIMIC database \citep{johnson2023mimic}. We note that ProbMED was pretrained on MIMIC using a different data split, such that its pretraining data may overlap with samples in our downstream evaluation split, potentially favoring its performance in this comparison. MedSigLip is the encoder of the MedGemma model family; it natively handles CXR views, as it was trained on de-identified medical data, including CXRs, dermatology images, ophthalmology images, and histopathology slides \citep{sellergren_medgemma_2026}. We finetune the MedSigLip encoder with frequency images of ECG signals such that it also has access to all three modalities during finetuning and testing.

\paragraph{Hospital-stay tasks.}
All hospital-stay tasks are defined per MIMIC-IV admission. \emph{In-hospital mortality} is a binary task that predicts whether the patient dies during the admission. \emph{ICD-10 Chapter~IX} (diseases of the circulatory system) is a multi-label task over the diagnosis codes assigned at discharge. We keep all Chapter~IX codes and code groups, down to a granularity of five characters, that occur in at least 1\% of hospital stays. \emph{DRG} comprises three multi-class tasks derived from the stay: the DRG code itself, the severity of illness and the risk of mortality. \emph{Length of stay} is a three-class task with the classes $<2$, 2--7 and $>7$ days. We use binary cross-entropy for the binary and multi-label tasks and cross-entropy for the multi-class tasks. We report the AUROC for in-hospital mortality and the macro-averaged (one-vs-rest) AUROC for all other tasks.  The finetuning pipeline for these tasks is illustrated in Fig. \ref{fig:finetuning}.

\paragraph{Unimodal tasks.}
\emph{CXR findings} is a multi-label classification of the 14 CheXpert labels. It is trained with binary cross-entropy, and we report the macro AUROC. \emph{LVEF} is regressed as a fraction in $[0,1]$ from the structured measurements that accompany the MIMIC-IV-Echo studies. The input is one apical four-chamber video per study, identified with the EchoPrime view classifier. \emph{ECG intervals} jointly regress six global intervals (RR, PR, P-wave duration, QRS duration, QT and QTc) from the 12-lead recording, using one output per interval. We derive the intervals from the machine measurements. Following \citet{Plagwitz2025QTcNetDeepLearning}, we discard records whose annotated intervals fall outside normal physiological bounds (PR $<20$ or $>500$\,ms; P wave $<20$ or $>300$\,ms; QRS $<50$ or $>300$\,ms; QT $<250$ or $>600$\,ms), as well as records carrying the "not measured" sentinel code, since these indicate lower annotation reliability. A record is retained only if all six targets are valid.

\paragraph{Coverage-filled metrics.}
Because the fine-tuned model uses only a fixed set of input modalities, it cannot make a prediction for evaluation samples that have none of these modalities (see \Cref{tab:modality_restriction}).
Metrics computed only on the covered samples (those with at least one required modality) can therefore look better than the model really is, and are not comparable across modality sets that cover different parts of the cohort.
To correct for this, we also report coverage-filled metrics, which are computed on every evaluation sample that has a valid label.
For each uncovered sample, the model's output is replaced by a fallback prediction that uses no input data and is estimated from the training split only.
For binary classification targets, the fallback is the training prevalence of the positive class, $\hat{p} = \frac{1}{N}\sum_{i} y_i$ (clipped to $[\epsilon, 1-\epsilon]$ with $\epsilon = 10^{-6}$). It is inserted as the logit $\log\frac{\hat{p}}{1-\hat{p}}$.
For multiclass targets, the fallback is the empirical class distribution $\hat{\boldsymbol{\pi}}$ from the training set, inserted as the log-probabilities $\log \hat{\boldsymbol{\pi}}$.
For regression targets, it is the mean of the training labels.
Formally, the filled prediction for sample $i$ is $\tilde{z}_i = m_i\, z_i + (1 - m_i)\, z^{\text{prior}}$, where $m_i \in \{0,1\}$ indicates whether any requested modality is available for sample $i$ and $z_i$ is the model output. We then compute the same metrics as in the covered-only evaluation on the filled predictions over the full labelled evaluation set: AUROC, average precision, accuracy and F1 for classification, and MAE and $R^2$ for regression.
A modality set with low coverage is thus penalized in proportion to the fraction of samples it cannot score, because those samples receive an uninformative, constant prediction. As a result, the coverage-filled metrics measure how useful a model is at the population level. As coverage approaches 100\%, they converge to the covered-only metrics.

\begin{figure}[t]
    \centering
    \includegraphics[width=0.85\linewidth]{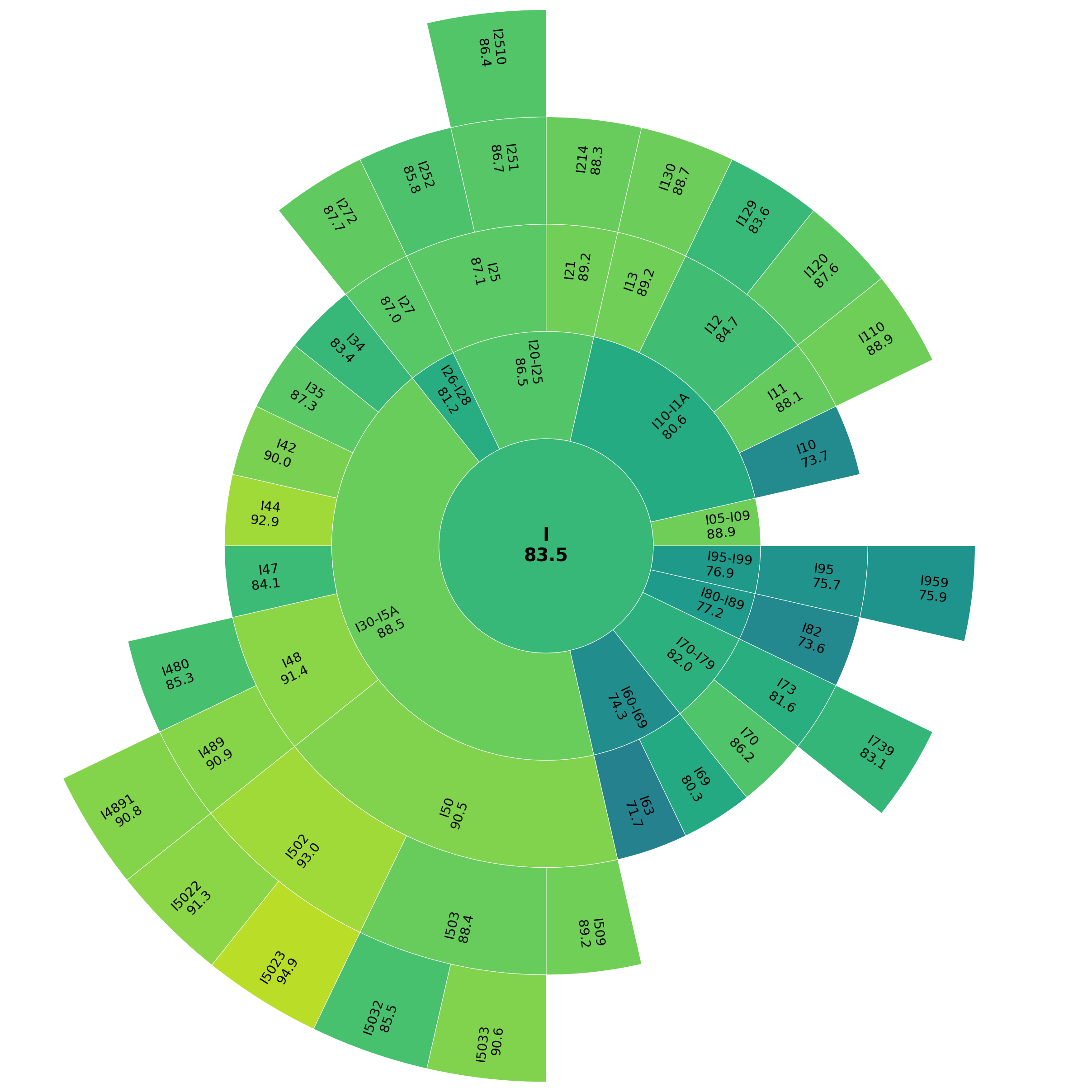}
    \caption{\textbf{ICD10 Ch. IX Classification Performance.}
    Finetuning on the complete MIMIC-IV dataset. Every cell represents an ICD10 code or a group thereof.
    The center cell is the root, whereas we reach the leaves towards the outer parts of the circle.
    The leaves are the most fine-grained codes.}
    \label{fig:icd10_radar}
\end{figure}

\begin{figure}[t]
    \centering
    \includegraphics[width=0.7\linewidth]{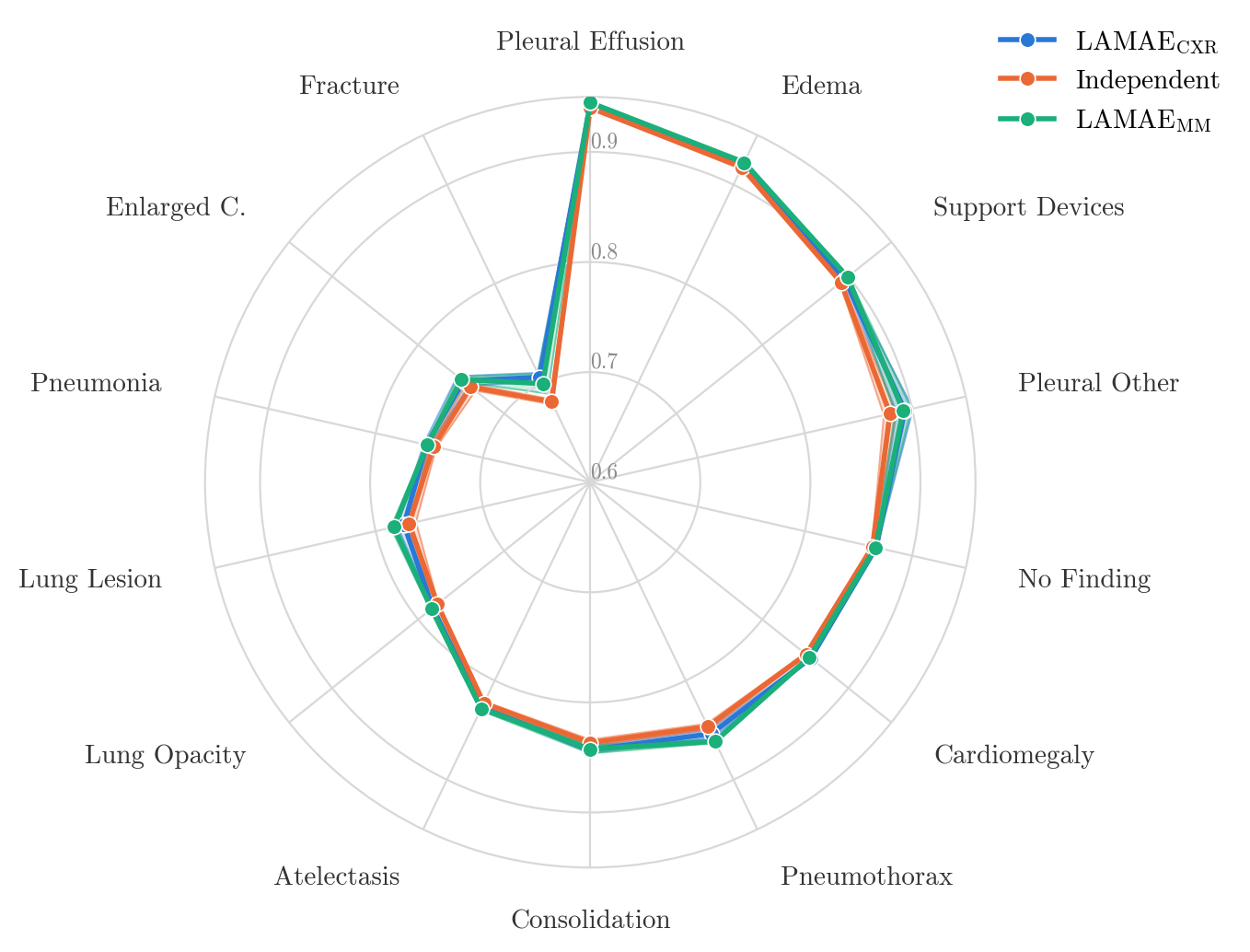}
    \caption{\textbf{Test AUROC for each of the 14 chest X-ray (CXR) labels}. Results obtained after fine-tuning LAMAE$_{\mathbf{\mathrm{CXR}}}$, Independent and LAMAE$_{\mathrm{MM}}$, all three using CXR input only at finetuning. Each point is the mean over three random seeds, and the shaded band shows standard deviation across seeds. Findings are sorted clockwise from the top by mean AUROC across the three models.}
    \label{fig:cxr_radar}
\end{figure}

\begin{figure}[t]
    \centering
    \includegraphics[width=0.7\linewidth]{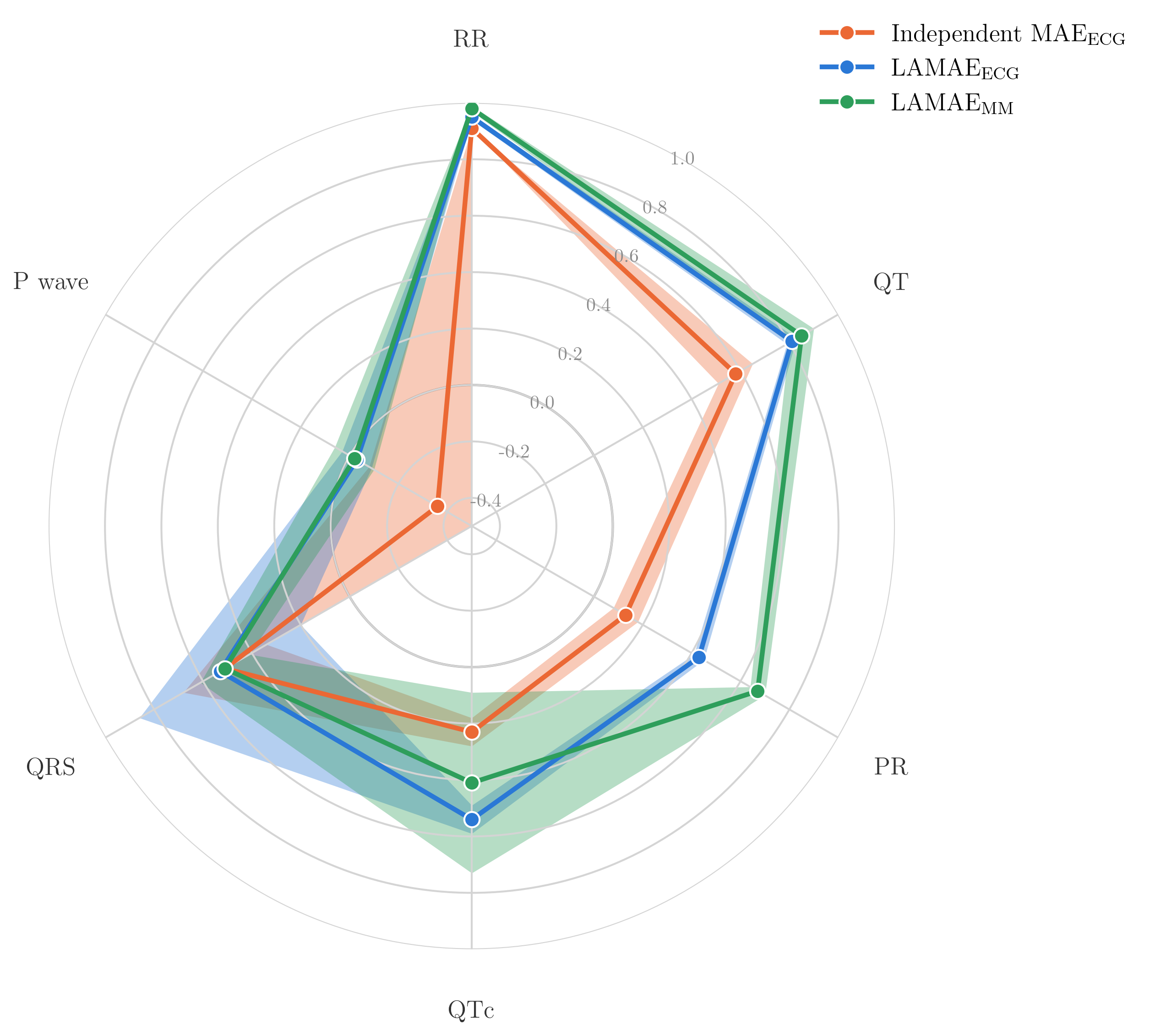}
    \caption{\textbf{ECG interval regression.} Test $R^2$ ($\uparrow$) per interval (mean $\pm$ std) over 3 seeds, Table~\ref{tab:unimodal_ecg}). }
    \label{fig:ecg_radar}
\end{figure}

\paragraph{Additional results.}
To complement the aggregate results reported in the main text, we provide per-label performance for the ICD-10 Chapter~IX and CXR classification tasks. \Cref{fig:icd10_radar} reports the test AUROC across individual ICD-10 Chapter~IX diagnosis codes and code groups, while \Cref{fig:cxr_radar} reports the test AUROC across the 14 CheXpert labels for LAMAE$_{\mathrm{CXR}}$, Independent MAE, and LAMAE$_{\mathrm{MM}}$. Figure~\ref{fig:ecg_radar} reports the test $R^2$ for each of the six ECG interval regression targets (RR, PR, P wave, QRS, QT and QTc) for Independent, LAMAE$_{\text{ECG}}$ and LAMAE$_{\text{MM}}$, as reported in Table~\ref{tab:unimodal_ecg}.


